# Static Seeding and Clustering of LSTM Embeddings to Learn from Loosely Time-Decoupled Events

Christian Manasseh

Razvan Veliche

Jared Bennett

Hamilton Clouse

## Abstract

*Humans learn from the occurrence of events in a different place and time to predict similar trajectories of events. We define Loosely Decoupled Timeseries (LDT) phenomena as two or more events that could happen in different places and across different timelines but share similarities in the nature of the event and the properties of the location. In this work we improve on the use of Recurring Neural Networks (RNN), in particular Long Short-Term Memory (LSTM) networks, to enable AI solutions that generate better timeseries predictions for LDT. We use similarity measures between timeseries based on the trends and introduce embeddings representing those trends. The embeddings represent properties of the event which, coupled with the LSTM structure, can be clustered to identify similar temporally unaligned events. In this paper, we explore methods of seeding a multivariate LSTM from time-invariant data related to the geophysical and demographic phenomena being modeled by the LSTM. We apply these methods on the timeseries data derived from the COVID-19 detected infection and death cases. We use publicly available socio-economic data to seed the LSTM models, creating embeddings, to determine whether such seeding improves case predictions. The embeddings produced by these LSTMs are clustered to identify best-matching candidates for forecasting an evolving timeseries. Applying this method, we show an improvement in 10-day moving average predictions of disease propagation at the US County level.*

## Acknowledgement

Work on this research has been funded by the Air Force Research Lab (AFRL) Autonomy Capability Team 3 (ACT3) under contract No. FA8649-20-C-0130.

## Introduction
In many real-life applications a dataset can consist of instances that have features that are both *static* and *dynamic*. For example, consider patient health data such as age and gender which are relatively *static* features compared to high-frequency *dynamic* heartbeat data as collected from electrode sensors connected to the patient. Sequence classification models such as Recurring Neural Networks (RNN) [1], Long Short-Term Memory (LSTM) [2], or Hidden Markov Models (HMM) [3] can be used to model the *dynamic* time-variant features of the event but are not suitable to address the static features [4]. In the patient health data example, an LSTM structure could be used to model the heartbeat timeseries data across multiple patients; however, it is not suitable for processing static and dynamic data simultaneously [4]. Ensemble methods such as those provided by Dietterich in [5] and Bagnall et al in [6] provide another way to address the issue: predictions made by temporal models such as Dynamic Time Warping (DTW) [7], Rotation Forests [8] and COTE [9] on dynamic data are combined with the predictions of a discriminative classifier on static data by performing distance measures as presented in [10]. Tzirakis et al. in [11] develop a methodology accomplishing simultaneously: (1) hierarchical clustering of raw dynamic data, (2) learning of deep end-to-end representations, and (3) temporal segments boundaries identification. They compute similarity between timeseries segments using an extension of DTW. A global loss function is used to optimize all three objectives. While this method results in representations learned from the clusters detected in this process, it does not intrinsically tie these representations to each timeseries.

In this work we introduce the concept of a Loosely Decoupled Timeseries (LDT) phenomenon and improve LSTM networks to enable Artificial Intelligence (AI) solutions to offer better timeseries predictions informed by static features or features that vary at a different frequency than the main event being modeled. A key feature of LSTMs is that they maintain a dual-purpose internal state (memory) that can aid in the learning and forecasting process [2]. We use similarity measures between timeseries based on the trends and introduce embeddings [12] representing those trends. The embeddings are constructed from features that are either static or changing at a different frequency than the timeseries of the main event being modeled. We apply this method to improve predictions of COVID-19 detected infections and death cases. We treat COVID-19 detected infections and death cases [13] as the main time-variant *dynamic* event and use socio-economic data at the US-County level as *static* features to inform predictions among counties of similar socio-economic structure but differing time lag in COVID-19 disease propagation among their population.

This work develops ideas from such disparate sources as COVID-19 forecasting, signature verification, and useful-life estimation from sensor data. Li et. al. [14] demonstrate improved signature verification by casting signatures as static representations of dynamic pseudo processes, using the dynamic process to generate an attention mechanism for the static representation. This has obvious ties to the COVID-19 pandemic as a dynamic process. We choose geospatial and demographic characteristics of communities as our static representation for two reasons: latent handling of mobility-impacted disease transmission and data augmentation. COVID-19 spread is heavily impacted by population mobility, which Panagopoulos et. al. [15] attempt to capture directly using graph neural networks, with vertices as cities and edges as movement between cities, while Xiao et. al. [16] use intra-city mobility patterns to train an adversarial encoder framework to predict next-at-risk communities. Both groups suffer from a lack of training data, which Panagopoulos et. al. [15] attempt to alleviate using transfer-learning between graphs generated from different countries. Wang et. al. [17] attempt to use augmented data for training, using an ABM to generate synthetic data based on an SEIR epidemic model. However, in our experience (unpublished), the SEIR model is not a great description of COVID-19 and the efficacy of epidemic models is highly dependent on their internal social-interaction model and estimated parameter values. Therefore, we propose a clustering approach based on geospatial and demographic attributes to augment our training data with other US counties that are similar in latent space and known-pandemic trajectory (matching COVID-19 spread based on where each county is in their respective pandemic trajectory)

Our novel contributions are as follows:

1) Definition and demonstration of Loosely-Decoupled Timeseries using static and trajectory-matched dynamic features for improved spatio-temporal prediction.
2) Computationally simple (K-means or K-medoids) latent-space clustering of static geospatial and demographic features accounting for mobility patterns and socioeconomic behaviors.
3) Built-in data augmentation through clustering of data with trajectory-matched pandemic behavior, effectively increasing the training data by reducing the number of prediction classes, thus obviating the need for potentially problematic synthetic data augmentation.

## Loosely Decoupled Timeseries

We define a Loosely Decoupled Timeseries (LDT) phenomenon as the relationship between two or more events that could happen at the same place or at different places but across different timelines sharing similarities in the nature of the event and the properties of the location. We contrast LDT with event-coupled timeseries [18] and tightly coupled timeseries [19]. Event-coupled timeseries consist of phenomena starting at the same time, whereas LDT allow for a lag between the event onsets. Tightly coupled timeseries start at the same time and are coupled in time throughout the event, such as the case of audio or speech and the corresponding video of lip gestures; whereas LDT events can happen at varying time frequencies such as the loose coupling of birth rates measured annually, and unemployment measured monthly. Other examples of LDT include the timeseries associated with a news cycle (hourly) in relation to the timeseries associated with the spread of violence (daily or weekly victim counts) or with the spread of disease (daily infection or death counts) being covered by the news cycle. LDT can also span two or more events occurring at different locations.

We represent LDT as:

$$\sim [\, x(t, env), x(a_1 * t + b_1, env_1), x(a_2 * t + b_2, env_2), \ldots ] \qquad (1)$$

where $x(t, env_i)$ is a time sequenced event [0, …., T] conditioned on the environment $env_i$, while $a_i$ provide for a varied frequency time series and $b_i$ provide for a time lag between the two events.

## Our Approach

A key feature of LSTMs is the maintenance of dual-purpose internal state (memory) that aids in learning and forecasting. This ameliorates the exploding or vanishing gradients problem experienced by RNNs [2], at the expense of slightly higher memory and computational complexity. This internal state convolves more distant and more recent information input; acting as a compression or embedding mechanism for the timeseries.

We use this internal memory state as an embedded representation of the timeseries, after appropriately training an LSTM model on subsequences of the timeseries.

We represent the LSTM model trained on timeseries $x(t, env_i)$, as [2]:

$$y^{out_j}(t) = f_{out_j}(net_{out_j}(t)); \quad y^{in_j}(t) = f_{in_j}(x_j(t, env_i))$$

where

$$net_{out_j}(t) = \sum_u w_{out_j u} y^u(t-1),$$

and

$$net_{in_j}(t) = \sum_u w_{in_j u} y^u(t-1).$$

We also have

$$net_{c_j}(t) = \sum_u w_{c_j u} y^u(t-1)$$

Which produces a trained LSTM model represented as:

$$L(T, env_i) \qquad (2)$$

The summation indices $u$, based on [2], can stand for input units, gate units, memory cells, or even conventional hidden units. These different types of units convey useful information about the current state of the LSTM. These may also be recurrent self-connections like $w_{c_j c_j}$. At time $t$, $c_j$'s output $y^{c_j}(t)$ is computed as:

$$y^{c_j}(t) = y^{out_j}(t) h(s_{c_j}(t))$$

The "internal state" or embedding representation $s_{c_j}(t)$ is:

$$s_{c_j}(0) = 0, \quad s_{c_j}(t) = s_{c_j}(t-1) + y^{in_j}(t) g\left(net_{c_j}(t)\right) \; for \; t > 0$$

The differentiable function $g$ "squashes" $net_{c_j}$; the differentiable function $h$ scales memory cell outputs computed from the internal state $s_{c_j}$.

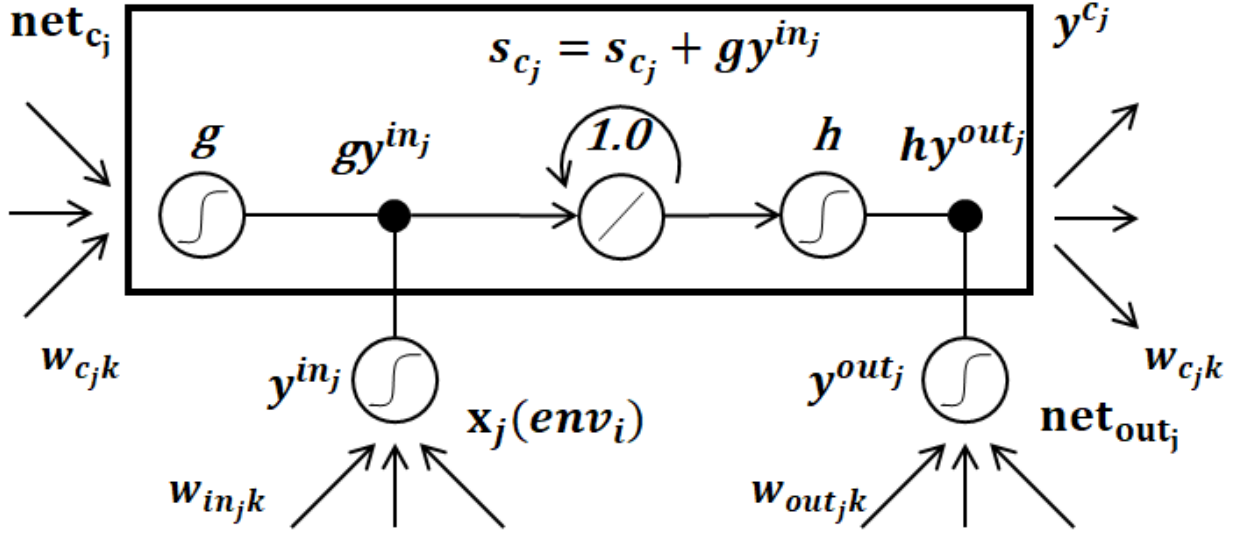

Figure 1. Architecture of memory cell $c_j$ (the box), the j-th memory cell block, and its gate units $in_j$, and $out_j$. The self-recurrent connection (with weight 1.0) indicates feedback with a delay of 1 time step. The index k ranges over hidden units u. [2]

We represent the embedding representation of $L(T, env_i)$ as:

$$s_c(T, env_i) \tag{3}$$

We enforce the same cell size, $u$, for all LSTMs trained to ensure the equidimensional representation of the internal states (3) to facilitate comparison independent of their length.

Our general focus is on the using the equidimensional embedding representations of (3) along with the time invariant LSTM-based prediction: given an evolving timeseries $x(t, env_i)$, search for and select longer/more evolved timeseries $x(a_j * t + b_j, env_j)$ that can be associated to produce a better prediction for the next time step(s) of (2) for $x(t, env_i)$.

To locate an associated timeseries, we expand the internal state (3) of an LSTM to include the *static* properties of $env_i$:

$$p_i(env_i) \tag{4}$$

We combine (3) and (4) to form:

$$[s_c(T, env_i), p_i(env_i)] \tag{5}$$

We call (5) the embedding representation of the phenomena at $env_i$. We then adopt a clustering method to cluster the embedding representations of several phenomena to identify the LDT tuples. In this paper we explore two clustering methods. K-means:

$$C_1, C_2, \ldots C_k = \text{argmin} \sum_{i=1}^{k} \sum_{x \in S_i} \|x - C_i\|^2 \tag{6}$$

With $k$ centroid points $C_k$ and minimizes the sum over each cluster of the sum of the square of the distance between the point and its centroid – the centroid is not necessarily a point from the data set.

And K-medoids:

$$M_1, M_2, \ldots M_k = \mathrm{argmin} \sum_{i=1}^{k} \sum_{x \in S_i} \|x - M_i\|^2 \tag{7}$$

which is similar to K-means but uses medoids $M_k$ chosen from points in the data set.

Once the LDT clustered tuples are identified, we then seed the main LSTM model based on this static data prior to each training episode, as well as prior to making a forecast. We call this seeding the local static embedding. This seeding takes the form of a dense embedding layer from the static data vector to the (initialization of the) hidden layer in the LSTM model. This dense embedding is trained by back propagation simultaneously with the LSTM training regimen. The embedding layer is reused at estimation time to reseed the hidden layer of the LSTM at the initial step.

# Forecasting COVID-19

COVID-19 is a disease (SARS-CoV-2) which spread rapidly around the world. Various metrics have been put forth to monitor its spread and evolution. While there is significant variation in the data collected, two main metrics of interest are generally available. One is the number of people "infected" (those that tested positive on a standardized testing platform), and the other is the number of people that died due to a recorded affiliation with the virus.

Both metrics are subject to reporting discrepancies. While some jurisdictions only report PCR-positive tests, the gold standard, others report results from less-reliable methods or even antibody tests (antibody tests rarely show positive until the end of an infection and then show positive for months or more after viral clearance). Additionally, death counts often represent both deaths due directly to COVID-19 complications and deaths due to unrelated causes but with positive test result (from preventative screening).

Simultaneously, there is a need from the public (and public health officials) to predict the evolution of these metrics days, potentially weeks, ahead for resource allocation and policy formulation. This prompted numerous efforts to build and apply predictive models [20] [21].

The standard models used in public health are derivatives of the SIR model [22]. These models are based on the "evolution" of an individual through the stages of a disease, from Susceptible "S" (has potential to get infected) to Infected "I" (virus is present) to Recovered "R" (disease ran its course). Variations considering asymptomatic, or unreported infections, as well as death as an outcome, are also used. While well understood, both from a theoretical and a practical (estimation) perspective, these models are necessarily limited by the assumption of compartmentalization (disease evolving in isolation). In reality, human movement patterns lead to diffusion of infection across boundaries. Solving coupled SIR compartmental models subject to constraints and diffusion becomes significantly more difficult and potentially intractable without deeper (longer history) samples.

Another aspect of compartmental models is their focus on inference rather than prediction. The primary focus of SIR-type models is on estimating disease characteristics (e.g., transmission rate) rather than prediction. Additionally, predictions themselves are only helpful up to a point. Just knowing something will happen is of limited usefulness, in the absence of scenario-based alternatives. From a prescriptive perspective (predictive + actionable) around the COVID-19 disease, it would be helpful to build upon similarities and "local tests" between US counties. By local tests we mean different restrictions and implementation or adherence to these restrictions, and their impact on the disease trajectory.

Working with researchers at the Air Force Research Lab / Autonomous Capabilities Team (ACT3), we apply the above-described methods of LDT-enhanced LSTM modeling on COVID-19 detected infection and death cases [13]. The time interval under study covered March through October 2020. During this time, the global events were such that there was a limited supply of COVID-19 testing resources, a hesitation in applying and adopting Non-Pharmaceutical Interventions (NPIs), and several US counties adopting lockdown procedures. During that same period, there was a single dominantCOVID-19 variant. From this perspective, the disease evolution was not influenced by mixed variants, as became more common in subsequent months. We limit the geographic span and resolution of our study to US County-level. County level data is more likely to contain a systematic definition of a COVID-19 case, following from locally consistent testing approaches and capabilities, consistent NPI measures, and consistent lockdown directives (if any were applied). More importantly, counties in the same state may experience a lag in disease spread among each other. Therefore, known data about the spread of the disease in counties already affected can inform the future state of counties starting to experience their first cases. In a similar fashion, the effect of NPIs (e.g., mask / lock-down mandates), observed in the infection and death timeseries of some counties, can inform the expected effect from similar NPIs in counties which are considering such measures. The value proposition of modeling county level data is thus significant from an operational and NPI implementation point of view.

Description of Data

The 2010 Census demographic datasets [23] for each of the US Counties were used in the analysis. Table 1 lists the data fields that were used in this study.

Table 1. 2010 Census demographic [23] data fields used in analysis.

| Data Field | Description |
| --- | --- |
| **SUMLEV** | Geographic Summary Level |
| **STATE** | State FIPS code |
| **COUNTY** | County FIPS code |
| **STNAME** | State Name |
| **CTYNAME** | County Name |
| **YEAR** | Year |
| **AGEGRP** | Age group |
| **TOT_POP** | Total population |
| **TOT_MALE** | Total male population |
| **TOT_FEMALE** | Total female population |
| **WA_MALE** | White alone male population |
| **WA_FEMALE** | White alone female population |
| **BA_MALE** | Black or African American alone male population |
| **BA_FEMALE** | Black or African American alone female population |

| Data Field | Description |
|---|---|
| AA_MALE | Asian alone male population |
| AA_FEMALE | Asian alone female population |
| TOM_MALE | Two or More Races male population |
| TOM_FEMALE | Two or More Races female population |

The economic data for each county was sourced from the USDA Economic Research Service [24] and Table 2 lists the data fields that were used in this analysis.

Table 2. USDA Economic Research Service [24] data fields used in analysis.

| Data Field | Description |
|---|---|
| FIPS_Code | State-county FIPS code |
| State | State abbreviation |
| Med_HH_Income_Percent_of_State_Total_2019 | County household median income as a percent of the State total median household income, 2019 |
| Median_Household_Income_2019 | Estimate of median household Income, 2019 |
| Unemployment_rate_2019 | Unemployment rate, 2019 |
| Unemployment_rate_2020 | Unemployment rate, 2020 |

In total, 385 socio-economic features from 3142 US Counties were used in the analysis. These constituted the static feature set and were assumed to remain constant during the analysis timeframe.

The COVID-19 infection metrics are aggregated by various entities. Johns Hopkins University [25] is an early and continuing resource for this data. However, they only collate what is reported by local health authorities, which are subject to local delays and constraints in identifying and reporting the disease spread.

For example, it has been observed that reported counts present a periodic dip around weekends. This is simply due to the limitations on scheduled activity for the labs running these tests. Correspondingly, there is a "bump" in counts at the beginning of the week, usually on Mondays.

The COVID-19 datasets from [13] [25] were used to analyze the daily number of infections and deaths. Table 3 lists the data fields that were used in the analysis.

Table 3. COVID-19 [13] [25] data fields used in the analysis.

| Data Field | Description |
|---|---|
| FIPS | State-county FIPS code |
| Date | (Implicit from file name) |
| Confirmed | Infections confirmed in area |
| Deaths | Deaths in area |

The number of cumulative infections and deaths was normalized against county population data. Raw data from [13] was used as is, with the exception of days in which a drop of cumulative infections or cumulative deaths were reported; during those days, the last reported value before the drop was used for all subsequent days until the cumulative values reached that level again.

LSTM Training Regimen

Focusing on prediction, rather than inference, potentially increases the utility of models from other domains. Time series analysis is one such domain, however, the structural constraints on those models are not easily aligned with a disease evolution expectation. An ideal model would "remember" trends and changes over varying time horizons (e.g., the convexity of the infected cases trend changed N days ago, where N could vary with the region under consideration). LSTM models have an established history in natural language processing, where learning the relationships between potentially distanced words help predict the next word. This is predicated on an underlying structure of the language from which samples are drawn (e.g., English), with long term memory keeping track of words earlier in the input. It is this long-term memory we had in mind when testing LSTM as a solution for predicting COVID-19 "trajectories". Given sufficient data, the model should learn to distinguish accelerating spread regions of the timeline from the more linear or saturated growth regions.

In our setup, the LSTM layer is followed by a dense layer with output dependent on the variables predicted. These variables are infected and dead counts, normalized by the population of the county.

LSTM training modules from PyTorch [26] were used and Ray Tune [27] was used to perform hyper-parameter tuning using grid-search. LSTM models were constructed to allow for grid-search across 64, 128, 256, and 512 hidden memory cells and across 1, 2, or 3 network layers. Several loss functions were tested for training: Mean Square Error (MSE), Relative Mean Square Error (RMSE), indexed or scaled versions to account for changing variability in the inputs across time, and versions penalizing for non-monotonic output. Input tensors covering a period between 7 and 30 days were used. Output was compared to actual values on 1, 3 and 5 day sliding intervals (days offset). The models with the top accuracy in prediction (lowest loss vs desired output) were successively retained by Ray Tune within the allocated time/computing resources; these models learned the most accurate representation for that county and point in time. Overall, more than 600 models were trained to extract county level embeddings over time.

For the purpose of consolidating our results, we consider two loss functions:

- MSE (abs): the absolute mean square error between the output and expected values
- RMSE (rel): the relative difference between output and expected values, with a large penalty imposed for producing non-monotonic sequences. A small quantity ($10^{-8}$) was added to the denominator to avoid dividing by 0.

The expectation was that RMSE based models will more closely match the disease trends, especially in the earlier stages when their population-normalized values are very small.

The following setup was used for all experiments:

- data for all counties covers the interval from first recorded case (for each US county) through 09/18/2020
- there is always a "buffer" of (the last) 30 days which are not "seen" by the trained models (test_days = 30); this means 09/19 through 10/18 is reserved for testing / evaluation of the model
- the models are trained for a certain time/computation "budget" using Ray Tune's ASHA Scheduler
- individual counties' training epochs consisted of one pass through all (chunked) historical data
- mini-batch training was used for all models (3 batches for individual counties models)

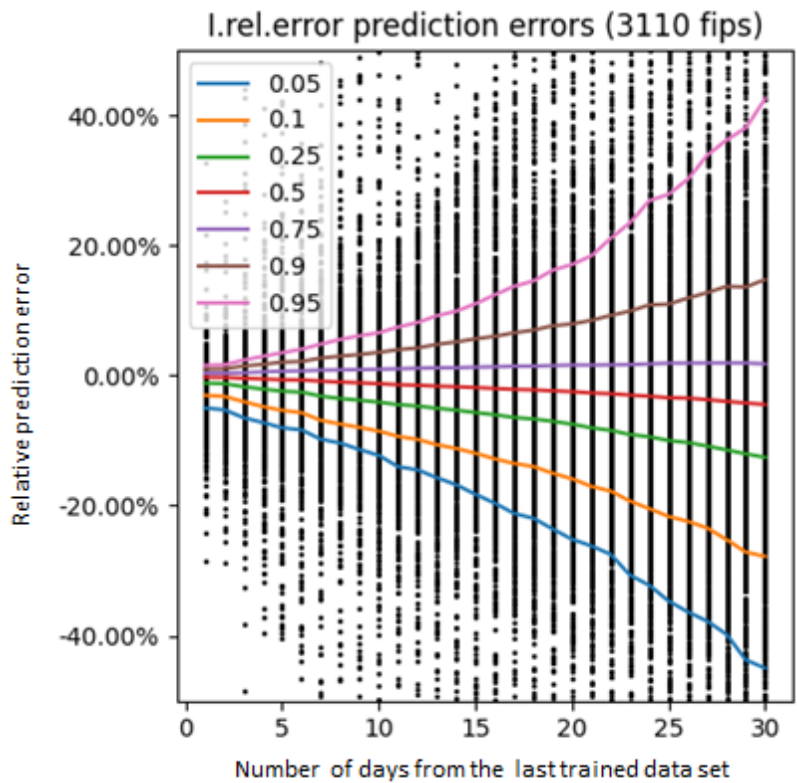

Figure 2. RMSE error plots for one US County showing future COVID-19 infection predictions for 1-30 days into the future. 5-day forward looking predictions are +/-3% accurate, and 10-day forward looking prediction errors are around +/-10% accurate.

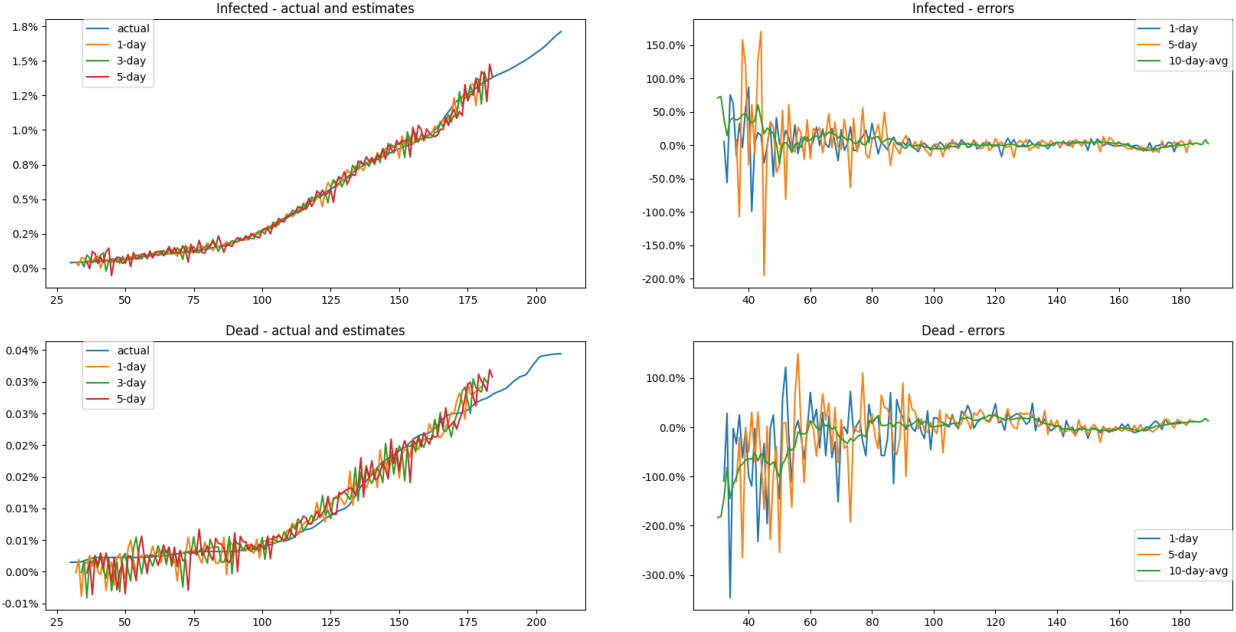

Figure 3. COVID-19 infection and dead 10-day-moving-average percent of population predictions contrasted against actual numbers for one US County. Predictions for 1, 3, and 5-forward looking days are provided on the left, and corresponding error rates (RMSE) are provided on the right. The x-axis in all plots represents days. The plots on the right show that, for this particular county, 90 and 120 days or more need to have passed from the onset of the disease in that county for the infection and death prediction error rates respectively to stabilize.

LSTM Hidden States as Embeddings

For each US County, the LSTM training regimen produces an optimum characteristic LSTM model for predicting number of infection and number of deaths for 1, 3 or 5 days in advance. The hidden state of each optimum trained LSTM model, representated as $L(T, env_i)$, with hidden state represented as $s_c(T, env_i)$ for County, 'i', was used to represent that US County. The hidden state was a vector of 256 dimensions representing the COVID-19 embedding for each US County, $[s_c(T, env_i), p_i(env_i)]$.

The US County embeddings were clustered using the two clustering methods in (6) and (7) with k = 3 clusters at 30, 60 and 90 day points in time. Clusters identified counties with similarity based on a shorter history, then matched other counties with a cluster, then analyzed the evolution of the groups' timeline to inform the new county's evolution.

Analysis of Results

The results presented in this section focus on analyzing 17 US Counties based in the state of Ohio with over 600 embeddings based on various points in time of the trained LSTM models for every county. The following notation is used in this section to represent the data being analyzed:

- Data is a vector of observed values over time and can be with or without socio-economic data
- Clustering method is either (6) or (7) with (6) represented with the label "k-means" and (7) as "k-medoids" in plots.

- Plot are taken at a Point-in-Time (PIT) reflected in the label of a plot and for a predefined set of clusters identified as cl:n where n is the number of clusters in the plot.
- Clustering involves all hidden layer states ("all") or only the last one ("last")

Figure 4 provides an example of k-means clustering into 3 clusters of 17 trained LSTM models with 60 days of training data using actual COVID-19 infection counts, relative to the total county population.

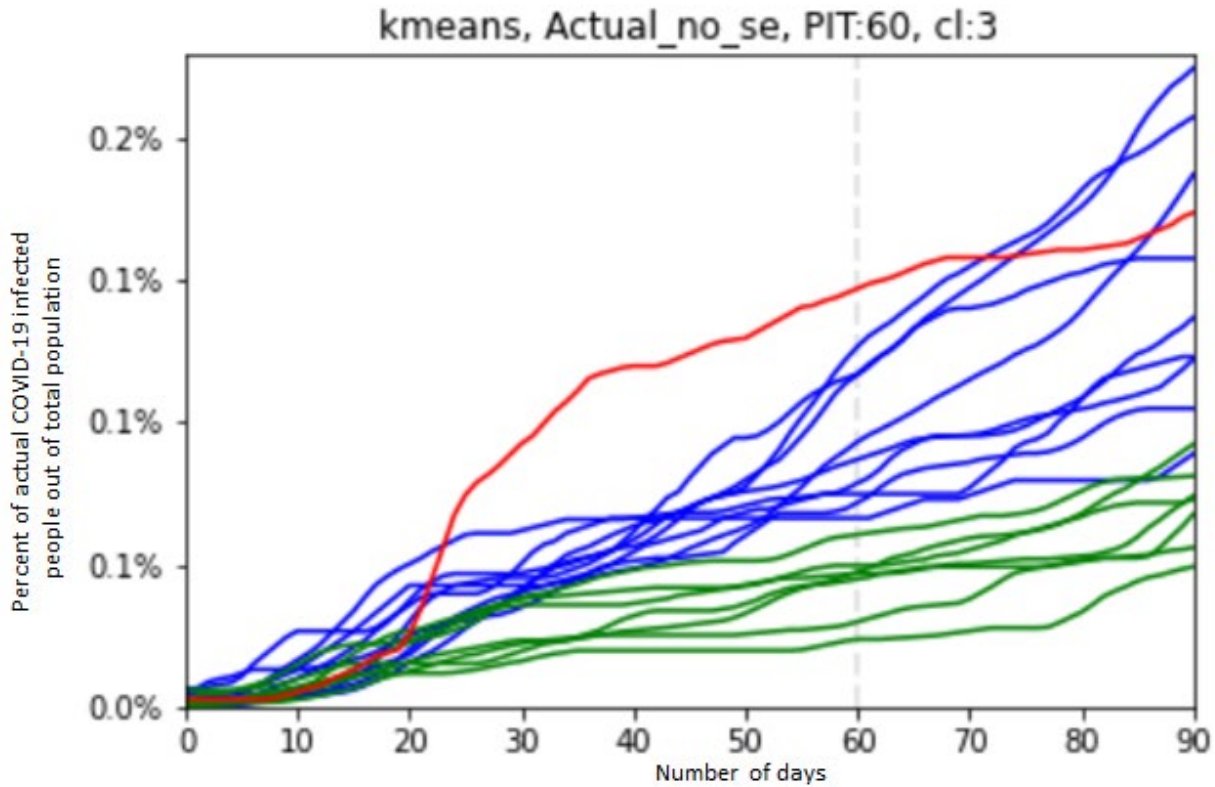

Figure 4. Example plot for 17 counties in OH clustered using k-means and the actual COVID-19 infections out of total population count. Socio-economic data was not factored in the clustering. Clustering was done using the hidden neural network layers (state) of an LSTM model trained with 60 days of data.

We analyze the alignment (concordance) of the two clustering methods (6) and (7) through two metrics:

1. Accuracy based on a confusion matrix
    a. Calculated as the percent of the diagonal values present in overall confusion matrix
    b. Dependent on the cluster order (based on a specific permutation of clusters)
    c. Has values ranging from 0 to 1, with 1 being more accurate
    d. Represented as "Acc" in plots
2. Adjusted Rand Index
    a. Takes into account the random chance "alignment" of clustering methods
    b. Independent of cluster order
    c. Has values ranging from -1 to 1, with 1 being more accurate.

d. Represented as "ARI" in plots

Figure 5 shows an example of Acc and ARI.

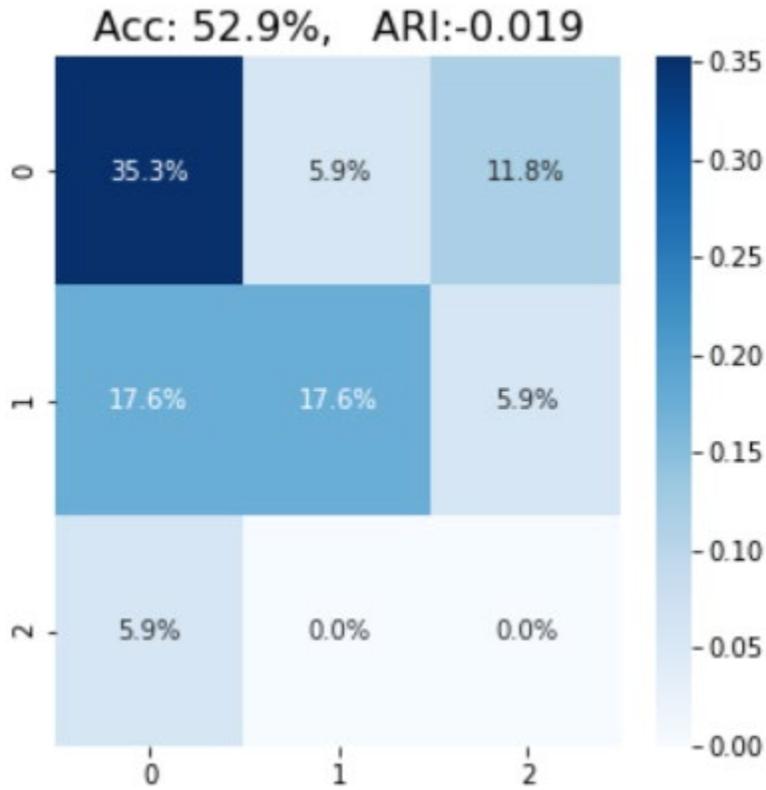

Figure 5. Example of Acc and ARI metrics derived from a confusion matrix of a clustering method, clustering data across 3 clusters.

The template representation depicted in Figure 6 shall be used to provide the analysis of the results.

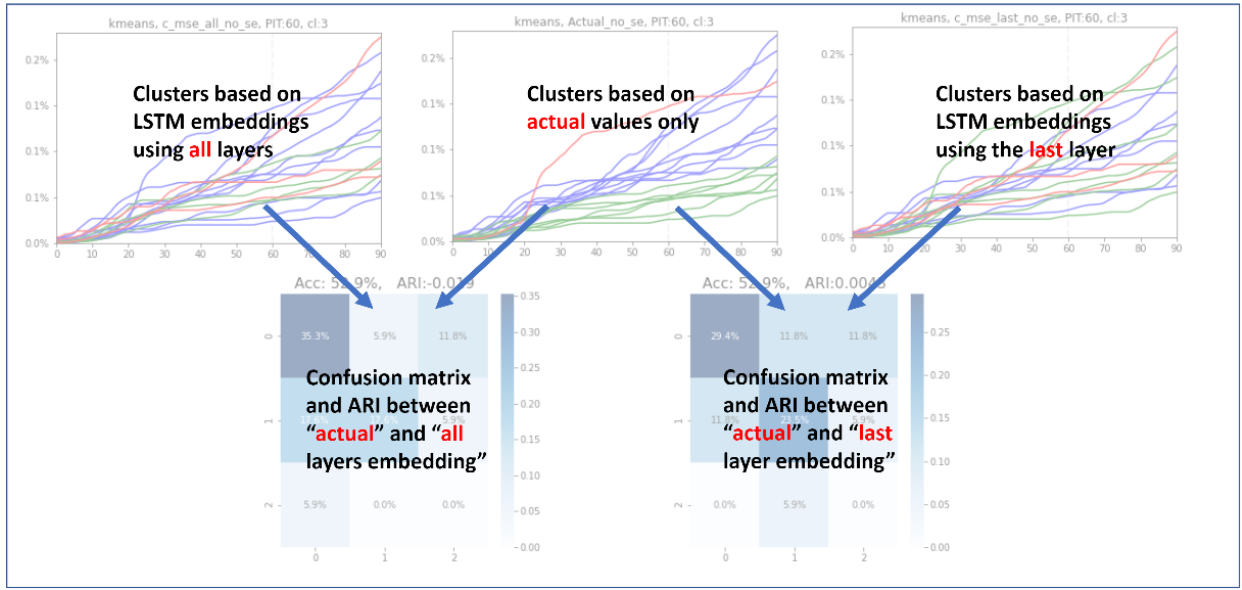

Figure 6. Template representation of results

The following 4 figures show the results of the analysis.

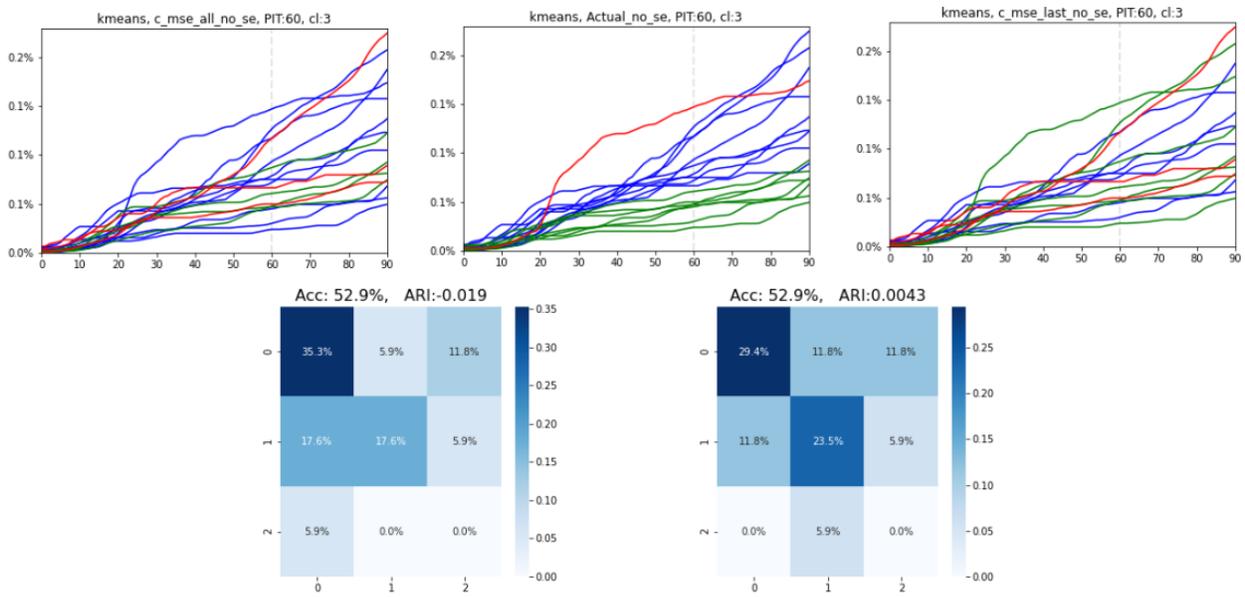

Figure 7. K-means with 3 clusters used to analyze PIT=60 with no socio-economic data.

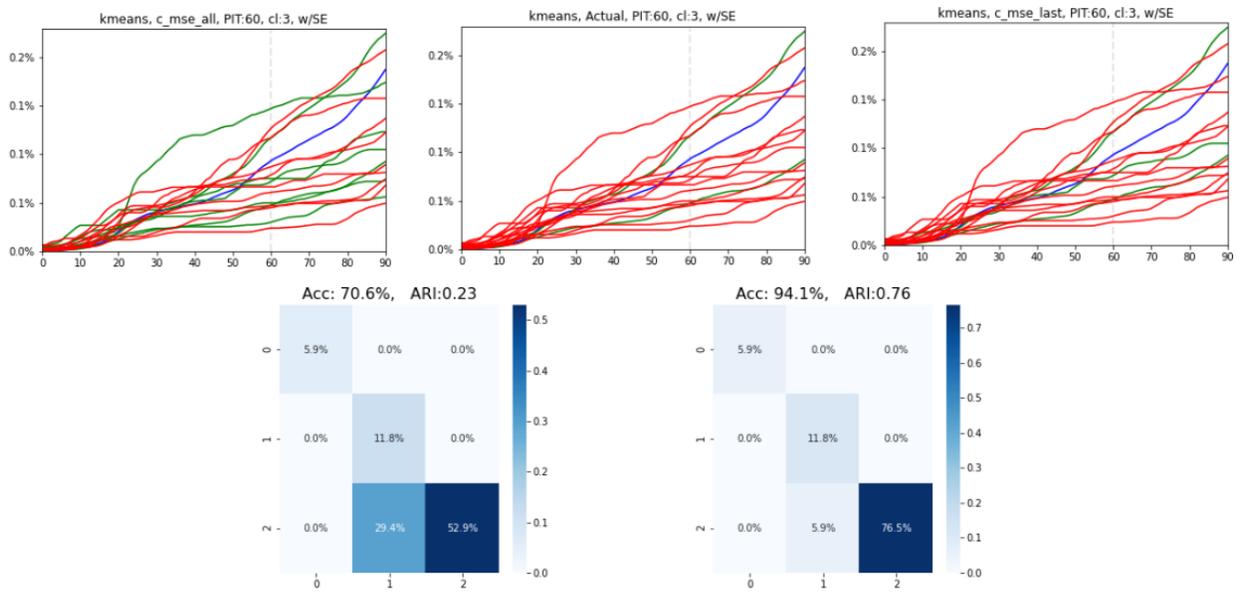

Figure 8. K-means with 3 clusters used to analyze PIT=60 with socio-economic data.

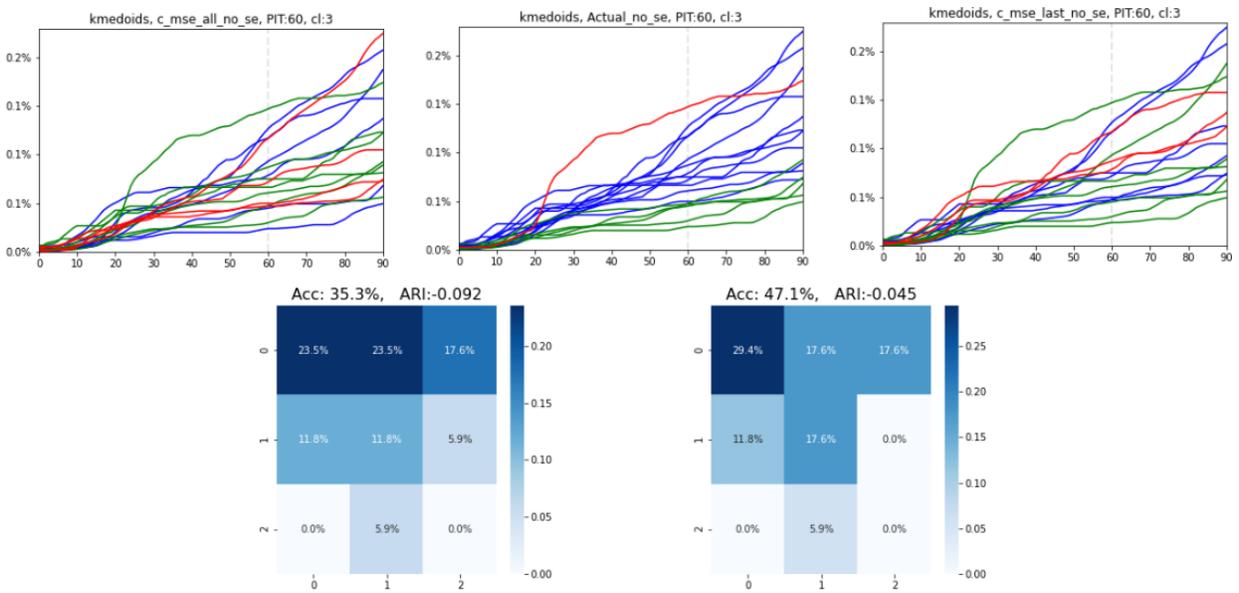

Figure 9. K-medoids with 3 clusters used to analyze PIT=60 with no socio-economic data.

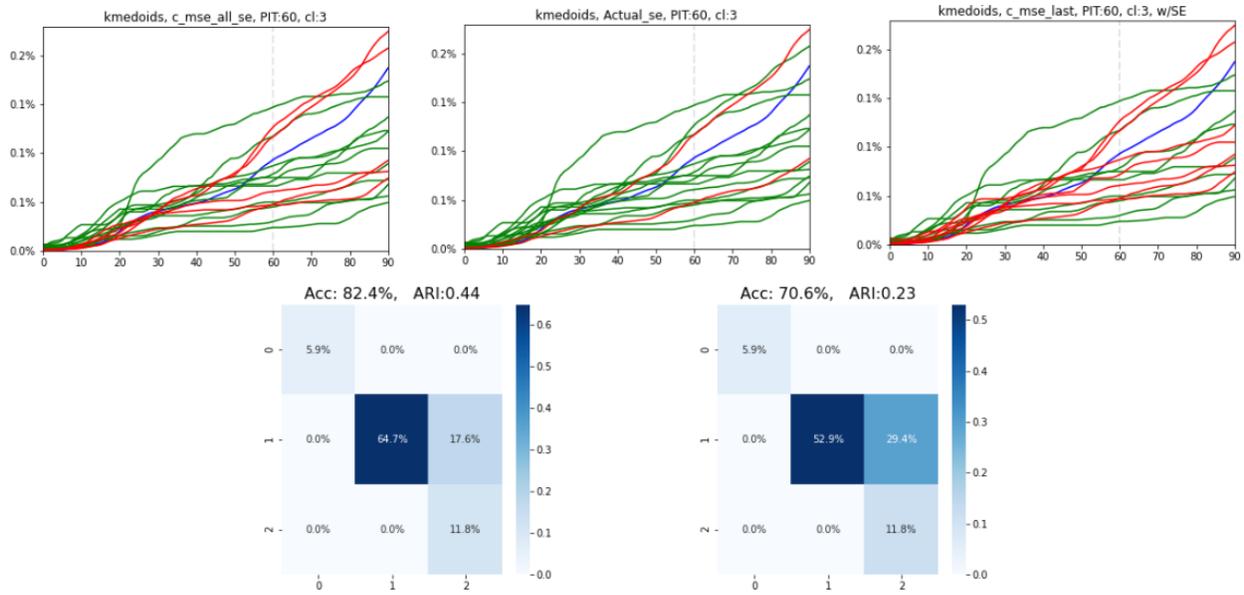

Figure 10. K-medoids with 3 clusters used to analyze PIT=60 with socio-economic data.

From Figure 7, Figure 8, Figure 9, and Figure 10 it can be concluded that including the socio-economic factors increases the consistency (overlap) between actual, observation-based clustering, and embedding-based clustering. It is also noticeable, that K-means (6) has an advantage over K-medoids and that using the last hidden layer of the LSTM as the embedding vector yields better results than using all the layers. The optimum clustering approach for COVID-19 data is to include the socio-economic data and use the last layer of the LSTM as the embedding in a k-means clustering algorithm.

We then proceeded to analyze the change in clustering over time. We define a "cluster stability" metric as follows (using Figure 11 as the example):

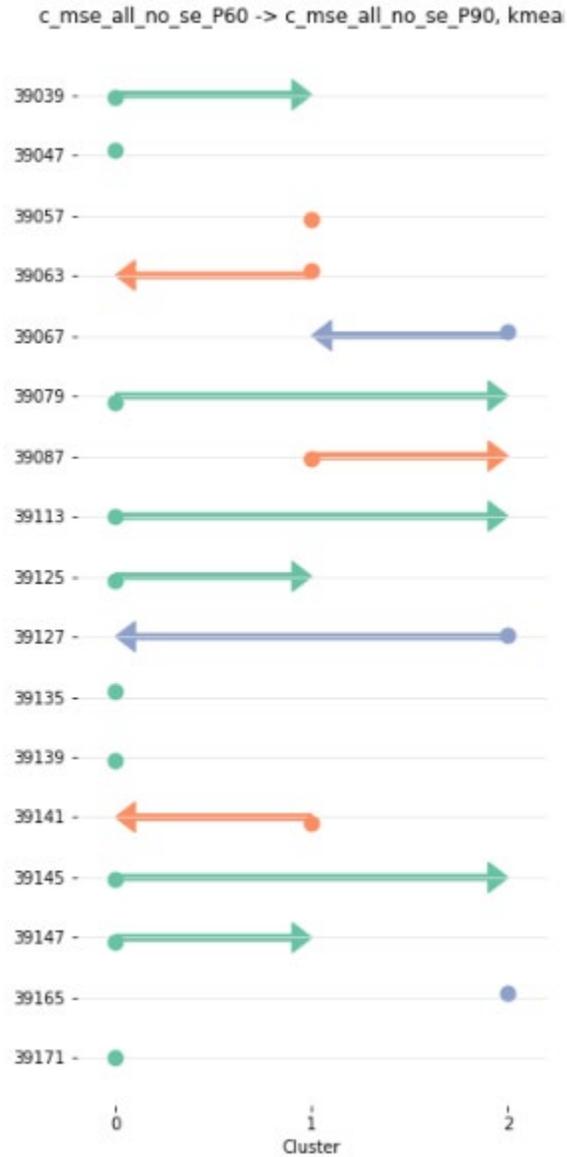

Figure 11. Using K-means and 3 clusters for the 17 counties in OH with no socio-economic data, this plot shows how the 17 counties changed cluster assignment between PIT=60 and PIT=90

- Cluster 0 (green dots) had initially 10 members
    - Of these, 3 shifted to cluster 1 and 3 shifted to cluster 2
- Cluster 1 (orange dots) had initially 4 members
    - Of these, 2 shifted to cluster 0 and 1 shifted to cluster 2
- Cluster 2 (blue dots) had initially 3 members
    - Of these, 1 shifted to cluster 0 and 1 shifted to cluster 1
- The overall *cluster stability* metric is defined as the maximal accuracy (for the optimal cluster reordering, yielding the maximum diagonal)
    - In this case: (4+1+1)/17 = 35.3%

Table 4. Cluster stability over time

| N (fixed out of 17) | All layers embedding | Actuals | Last layer embedding | All layers error | Last layer error |
|---|---|---|---|---|---|
| Without SE factors | Acc = 35.5% N = 6 | Acc = 76.5% N = 13 | Acc: 41.2% N = 7 | \|13-6\|/13 = 54% | \|13-7\|/13 = 46% |
| With SE factors | Acc: 58.8% N = 10 | Acc: 100% N = 17 | Acc: 58.8% N = 10 | \|10-17\|/17 = 41% | \|10-17\|/17 = 41% |

Table 4 shows the cluster stability calculations (last two columns) across two types of embeddings (rows): without socio-economic data and with socio-economic data. The two columns reflect the calculations made using LSTM embeddings derived from the last layer or all layers of the neural network.

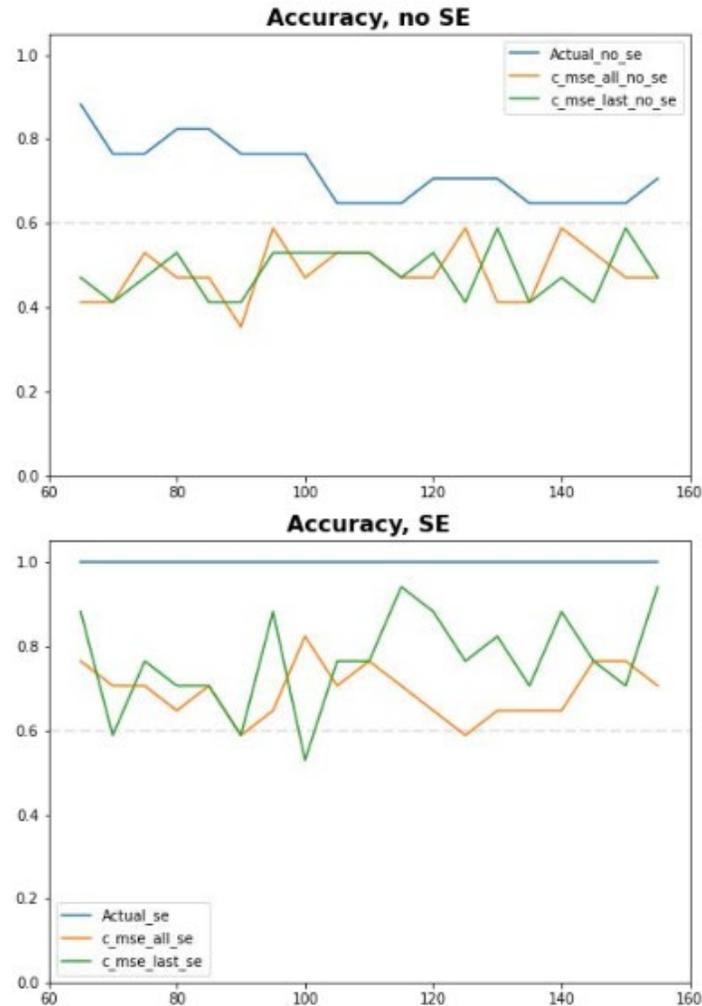

Figure 12. Cluster Stability plotted over time in comparison with PIT=60 for the cases of (top) without socio-economic data and (bottom) with socio-economic data.

Figure 12 plots the cluster stability metrics over a wide range of PIT variations and compares the effect of socio-economic data on the stability of the clusters. Adding the socio-economic data to the LSTM embeddings stabilized the cluster formation over time.

## Conclusion

The work presented here demonstrates improved LSTM forecasting through embeddings derived from loosely-decoupled timeseries. We applied this methodology to COVID-19 infection and death at the US county level. Our socio-economic embedding approach demonstrates enhanced 10-day moving average predictions compared to traditional LSTM modeling, especially in conjunction with K-means clustering of the final layer embeddings. Additionally, we demonstrate stability in the clustering of LDTs when combined with socio-economic data, providing increased consistency in predictions. With this approach, US counties later in catching the virus benefit from counties similar in socio-economic demographics, but with an earlier start to their disease propagation, improving the predictive outcome.